\documentclass[letterpaper, 10pt, conference]{ieeeconf}
\IEEEoverridecommandlockouts
\usepackage{mathptmx}
\usepackage{times}
\usepackage{amsmath}
\usepackage{amssymb}
\usepackage{dsfont}
\usepackage[cal=cm]{mathalfa}
\usepackage{esvect}

\usepackage{graphicx}
\usepackage{subcaption}
\usepackage{overpic}
\usepackage{wrapfig}
\usepackage{tikz}
\usetikzlibrary{calc,spy}
\usepackage[export]{adjustbox}

\usepackage{aprl_acronyms}
\usepackage{aprl_misc}

\usepackage{array}
\usepackage{tabularx}
\usepackage{booktabs}
\usepackage{makecell}
\usepackage{multirow}

\usepackage[table]{xcolor}
\usepackage{soul}

\usepackage{bbding}
\usepackage{pifont}
\usepackage{wasysym}

\usepackage[ruled,linesnumbered]{algorithm2e}

\usepackage{lipsum}
\usepackage{stackengine}

\usepackage{cite}
\makeatletter
\let\NAT@parse\undefined
\makeatother
\usepackage[numbers,sort&compress]{natbib}

\PassOptionsToPackage{hyphens}{url}
\usepackage{xurl}
\usepackage[breaklinks,colorlinks,citecolor=black,linkcolor=black]{hyperref}
\usepackage[capitalize]{cleveref}

\title{\LARGE \bf
LT-Mem: Volatility-Aware Spatio-Temporal Memory \\
for Lifelong Scene Understanding}

\author{
    Yumin Lee$^{1}$, Hyoseok Ju$^1$, and Giseop Kim$^{1*}$%
    \thanks{$^{*}$Corresponding author.}%
    \thanks{$^{1}$Y. Lee, H. Ju and G. Kim are with the Department of Robotics and Mechatronics Engineering, DGIST, Daegu, Republic of Korea {\tt\small [yumin.lee, hyoseokju, gsk]@dgist.ac.kr}}%
    \thanks{This work was supported by the National Research Foundation of Korea (NRF) grant funded by the Korea government (MSIT) (RS-2026-25492530), by Basic Science Research Program through the National Research Foundation of Korea (NRF) funded by the Ministry of Education (No. RS-2025-25420118), and by the Institute of Information \& Communications Technology Planning \& Evaluation (IITP) grant funded by the Korea government (MSIT) (No. RS-2025-02219277, AI Star Fellowship Support (DGIST)).}
}

\begin{document}
\maketitle
\thispagestyle{empty}
\pagestyle{empty}


\begin{abstract}
Long-term robot operation in evolving environments requires object-level understanding that persists across repeated revisits.
Existing systems either overwrite history to maintain an up-to-date map or store semantic snapshots without consistent cross-session object identity, resulting in \emph{temporal amnesia}: the systematic loss of object history that prevents answering queries such as ``Where has the green chair been across all sessions?''
We propose LT-Mem, a volatility-aware memory evolution framework that unifies spatially aligned instance-level 3D perception with volatility-conditioned temporal reasoning.
First, a multi-session SLAM backbone provides spatially aligned per-object observations across sessions.
Second, a reasoning layer governs how object memory evolves: deterministic evidence scoring preserves cross-session identity, and a volatility-aware policy selects among overwrite, hold, and multi-hypothesis actions based on each object's dynamics.
Third, the resulting Tri-Memory structure (Live, Delta, Meta) preserves both current states and event histories, enabling longitudinal object-centric reasoning.
We further introduce LT-VQA, a dataset and evaluation suite comprising multi-session recordings, persistent identity annotations, and temporal QA pairs.
Experiments show that LT-Mem consistently outperforms baselines across all metrics while consuming an order of magnitude fewer tokens, and ablations confirm that gains are driven by the structured memory architecture rather than LLM capacity.
\end{abstract}

\section{Introduction}
\begin{figure}[t!]
  \centering
  \includegraphics[width=\linewidth,keepaspectratio]{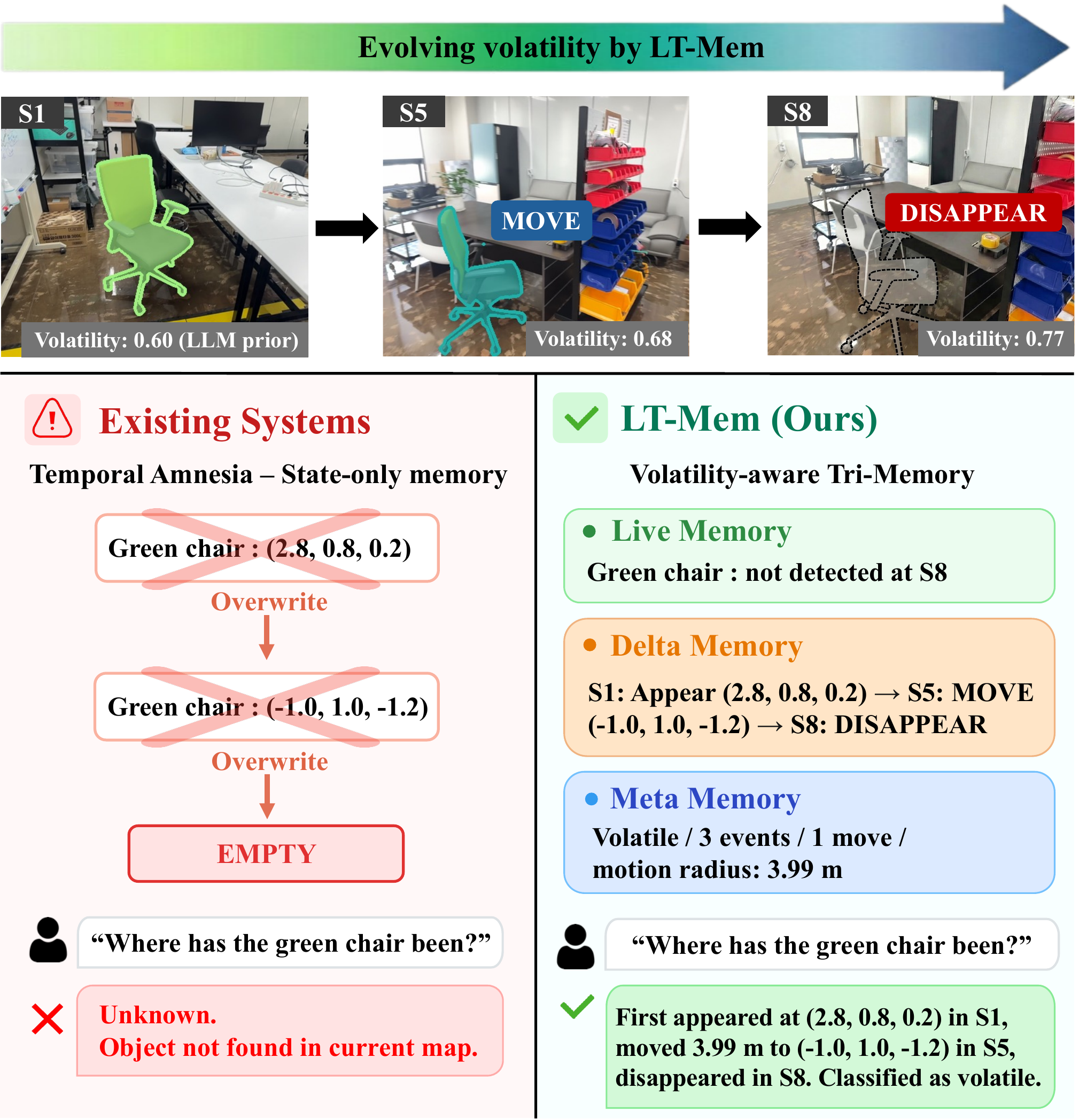}
  \caption{
  Overview of the proposed volatility-aware Tri-Memory framework
  for multi-session environments.
  Object observations across revisits are interpreted as events
  and integrated into Live, Delta, and Meta memories.
  The framework adapts update decisions to object volatility,
  enabling longitudinal object-centric reasoning and visual question answering.
  }
  \label{fig:framework-overview}
\end{figure}

%

%
\begin{table*}[t!]
\centering
\caption{Taxonomy of object-level events and corresponding longitudinal queries supported by LT-Mem. The upper section defines event types with representative queries; the lower section lists compound query types that require reasoning across multiple sessions.}
\label{tab:transition_taxonomy}
\resizebox{\textwidth}{!}{%
\begin{tabular}{llll}
\toprule
\textbf{Event} & \textbf{Description} & \textbf{Example Query} & \textbf{Required Memory} \\
\midrule
\textsc{Appear}   & Object first observed in session $S_t$      & ``When did the scissors first appear?''              & Delta \\
\textsc{Disappear} & Object no longer detected after $S_t$       & ``Was the vacuum in the room at Session~9?''            & Delta \\
\textsc{Move}     & Centroid displacement beyond threshold       & ``Did the brown basket move at Session~6?''    & Delta\\
\textsc{None}   & No state change detected in session $S_t$       & ``Did the fridge stay in place at Session~4?''               & Delta \\
\textsc{Re-appear} & Object returns after prior disappearance    & ``Did the green chair come back at Session~9?''                       & Delta \\
\midrule
\multicolumn{4}{l}{\textbf{Compound (multi-session reasoning)}} \\
\midrule
Trajectory           & Sequence of transitions across all sessions      & ``Where has the  blue tote bag been across all sessions?'' & Delta, Meta, Live \\
Volatility           & Frequency and magnitude of changes           & ``Which object moved most frequently?''               & Meta \\
Temporal localization & Identifying \textit{when} a change occurred & ``When was the last time the white board moved?''         & Delta \\  
Counterfactual       & Comparing states at two specific sessions    & ``Was the robot dog in the same location in $S_1$ and $S_{10}$?'' & Delta, Live \\
\bottomrule
\end{tabular}%
}

\vspace{-5mm}

\end{table*}

Robots repeatedly revisit evolving environments
\cite{carlone2025slam, yin2025general, cadena2017past}. Across sessions,
objects may move, disappear, or be inconsistently observed due
to occlusion and alignment noise \cite{schmid2024khronos}.
Multi-session mapping provides a shared spatial frame
\cite{pomerleau2014long, kim2022lt}, but spatial consistency alone
does not guarantee that object memory remains valid over time.
Lifelong operation requires reasoning about \emph{when} and \emph{how} object
states should be updated, preserving trajectories of change
rather than collapsing them into a single overwritten state.
For example, answering a session-indexed query such as:
``At Session~3, the robot dog was near the white desk.
Where is it at Session~9?'' requires persistent cross-session
identity and explicit modeling of state transitions.
Current systems are not designed for this form of
spatio-temporal object-centric reasoning.

Existing lifelong scene understanding approaches follow two
directions: geometry-centric mapping and semantic memory.
Geometry-centric systems
\cite{pomerleau2014long, schmid2024khronos, kim2022lt}
maintain global consistency under change, but reduce object
transitions to deletion and re-creation, losing temporal
history. Semantic memory approaches
\cite{anwar2025remembr, gu2024conceptgraphs, yang20253d}
construct rich representations via view-level embeddings,
open-vocabulary 3D scene graphs, or structured exploration
memories. However, they typically rely on the latest
observation or treat sessions independently, without enforcing
persistent cross-session identity. Even when long-horizon
context is incorporated, reprocessing accumulated historical
video streams becomes computationally demanding and does not
scale with prolonged deployment. Consequently, object
histories cannot be coherently composed, and session-indexed
temporal queries are not directly supported.

To bridge these directions, we first analyze the taxonomy of
cross-session object state transitions and the longitudinal
queries they entail, as summarized in
Table~\ref{tab:transition_taxonomy}. This taxonomy clarifies
the types of memory evolution required for sustained
long-term operation. Building upon this analysis, we propose
\textbf{LT-Mem}, a volatility-aware memory evolution framework
with two layers: a perception layer for spatially aligned
object observations, and a reasoning layer that governs
memory updates.
\begin{itemize}
\item \textbf{Perception layer.}
We extend MASt3R-SLAM~\cite{murai2025mast3r} for
multi-session alignment and apply instance-level 3D
segmentation~\cite{carion2025sam} to extract
per-object centroids, volumes, and semantic embeddings.

\item \textbf{Reasoning layer.}
Cross-session identity is resolved through deterministic
evidence scoring (E1--E5 in \tabref{tab:evidence}), including appearance and
spatio-temporal consistency, with a constrained LLM used only
for ambiguous cases. A volatility-aware policy then selects
among overwrite, hold, and multi-hypothesis updates based on
object dynamics and alignment quality.
\end{itemize}

The reasoning outputs are organized into a
\textbf{Tri-Memory structure}: Live Memory stores current
states, Delta Memory records timestamped events, and Meta
Memory accumulates long-term statistics such as volatility.
This preserves both present states and historical trajectories,
enabling long-horizon temporal queries.
Because the reasoning layer operates over compact structured
representations rather than raw visual streams, LT-Mem achieves strong temporal reasoning with 
per-session token cost independent of frame count, 
enabling scalable lifelong operation.

To evaluate this capability, we introduce the
\textbf{LT-VQA} dataset\footnote{\url{https://lt-mem.github.io/}},
which provides multi-session recordings with aligned poses,
per-object annotations, and temporal QA pairs. Unlike
single-session QA datasets such as NaVQA~\cite{anwar2025remembr},
LT-VQA targets long-term map management across sessions,
evaluating whether the system correctly identifies what
changed, when, and across which sessions.

In summary, this work makes the following contributions:

\begin{itemize}
\item We propose LT-Mem, a volatility-aware memory evolution
framework in which update decisions are governed by a
volatility-conditioned reasoning layer, optionally assisted by
a constrained LLM, producing a Tri-Memory structure
(Live, Delta, Meta) that preserves both current and historical
object states.

\item We construct LT-VQA, a dataset and evaluation suite requiring persistent
object identity and structured temporal reasoning across
multi-session revisits.

\item We demonstrate that volatility-aware evolution improves
long-horizon object-centric reasoning over overwrite and accumulation baselines, while reducing token consumption by up to an order of magnitude compared to vision-based approaches.
\end{itemize}

\section{Related Work}
\label{sec:related-work}

\subsection{Lifelong Spatial Mapping}
\label{subsec:spatial-mapping}

Long-term autonomy requires maintaining spatial consistency across repeated revisits while accommodating environmental change. Classical systems such as LT-mapper~\cite{kim2022lt} and \citeauthor{pomerleau2014long}~\cite{pomerleau2014long} detect map--observation discrepancies and resolve them by removing or correcting outdated elements. More recently, Khronos~\cite{schmid2024khronos} constructs dense spatio-temporal metric-semantic maps that unify short-term dynamics and long-term scene changes, but does not maintain an identity-preserving event log for individual objects. Even probabilistic approaches that track per-object stationarity scores~\cite{qian2022pocd} ultimately discard past object states to refresh the map rather than retaining state histories. When an object relocates from position~A to~B, the prior state is overwritten or removed rather than recorded as part of a temporal trajectory. We refer to this systematic loss of object state history as \emph{temporal amnesia}, a limitation that prevents answering longitudinal queries such as ``How many times has the robot dog moved?''

\begin{figure*}[t]
  \centering
  \includegraphics[width=0.9\textwidth]{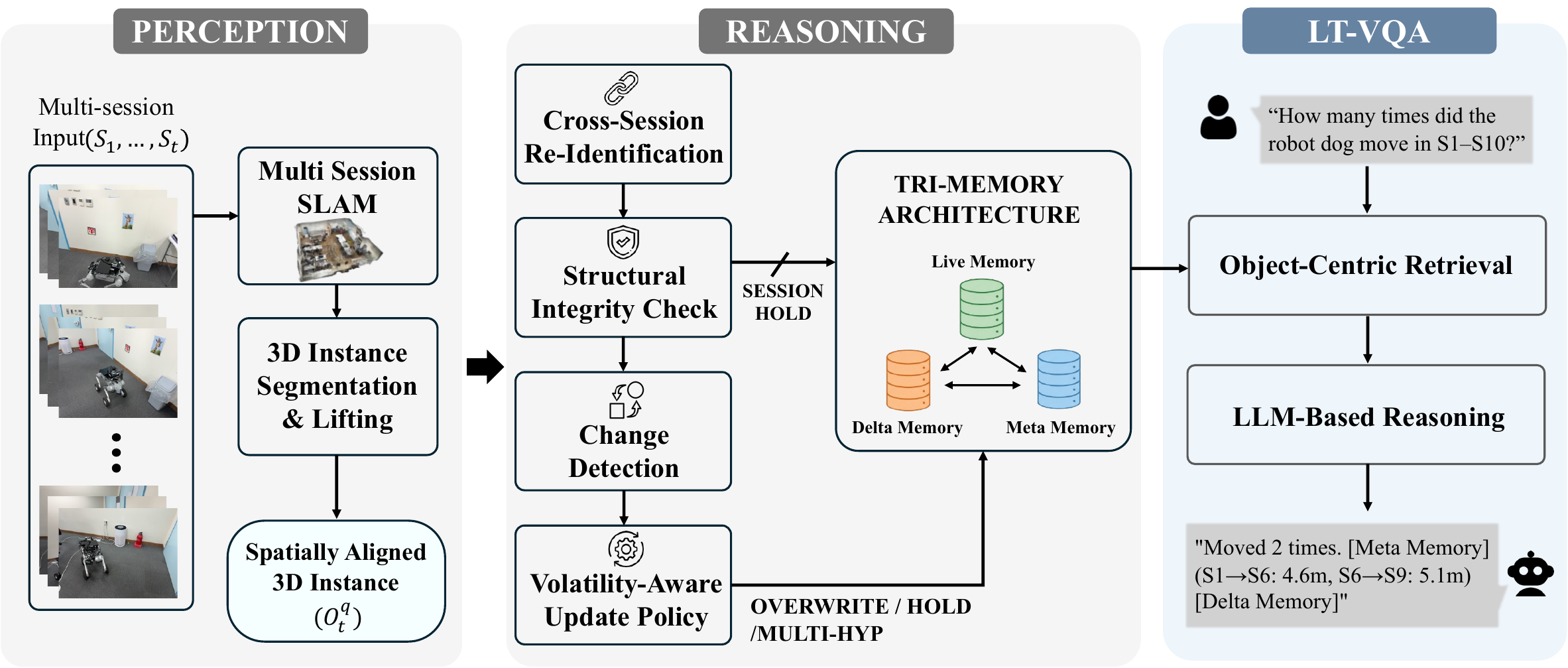}
  \caption{System pipeline. (Left) The Perception layer aligns
  multi-session observations and extracts object instances via 3D
  segmentation. The Reasoning layer performs integrity-checked memory evolution, including identity matching, change detection, and volatility-aware update selection. (Right) Updated states are
  organized in the Tri-Memory (Live, Delta, Meta) and retrieved
  through object-centric retrieval to support long-term visual question answering.}
  \label{fig:pipeline}
  \vspace{-5mm}
\end{figure*}

\subsection{Semantic Scene Memory}
\label{subsec:semantic-memory}

A complementary direction constructs a robot's long-term memory around semantic representations. View-based systems such as ReMEmbR~\cite{anwar2025remembr} accumulate observation-level embeddings for retrieval-augmented reasoning, while object-level approaches embed foundation model features into 3D representations such as point clouds~\cite{gu2024conceptgraphs} or scene memory~\cite{yang20253d}.  A recent approach~\cite{gorlo2025describe} further extends this to a hierarchical 4D scene graph with detailed spatio-temporal descriptions.
However, view-level memory does not enforce explicit cross-session data association; two observations of a ``chair'' may or may not correspond to the same physical instance. Object-centric semantic maps operate within a single session and typically commit to the most recent state without modeling transitions, exhibiting the same temporal amnesia as geometry-centric systems. Even retrieval agents that recall raw observations on demand~\cite{chen2026searchingspacetimeunified} keep only raw records, so neither family maintains the \emph{structured, identity-resolved temporal histories} required for longitudinal object-centric reasoning.

\subsection{3D Foundation Models and the Perception--Reasoning Gap}
\label{subsec:perception-reasoning-gap}

Dense reconstruction methods such as DUSt3R~\cite{wang2024dust3r} and MASt3R~\cite{murai2025mast3r} enable metrically accurate multi-view geometry without task-specific training, while universal segmentation models such as SAM3~\cite{carion2025sam} provide instance-level object extraction across diverse environments. Yet enhanced perceptual accuracy alone does not resolve the long-standing update dilemma: whether to overwrite or accumulate conflicting observations. LT-Mem bridges this gap by coupling spatially grounded object instances with a volatility-conditioned reasoning layer that governs identity and update logic, producing the Tri-Memory structure as a natural organization of its heterogeneous outputs.

\section{Method}

\subsection{Problem Formulation}

We consider an environment observed across $T$ sessions (i.e., independent mapping runs conducted at different times) $S_1, \ldots, S_T$, where objects may move, appear, disappear,
or be temporarily occluded.
For each object $q$ in session $S_t$, the perception layer produces a
3D observation

\begin{equation}
  o_t^q = \left(\mathbf{c}_t^q,\; v_t^q,\; \mathbf{f}_t^q\right),
\end{equation}
where $\mathbf{c}_t^q \in \mathbb{R}^3$ is the centroid,
$v_t^q$ denotes volume,
and $\mathbf{f}_t^q$ is a visual embedding extracted from the segmented region.
The system maintains a persistent memory $\mathcal{M}^q$
that is updated after each session to support long-term temporal
queries such as
\textit{``Where is the white chair now?''}
and
\textit{``How often does the robot dog move?''}

\subsection{Perception Layer}
\subsubsection{Multi-Session \ac{SLAM}}

We build on MASt3R-SLAM~\cite{murai2025mast3r},
which estimates camera poses and dense point maps from monocular video
using 3D reconstruction priors.
To align observations across sessions into a shared global frame,
we build on MR.ScaleMaster~\cite{ju2026mrscalemasterscaleconsistentcollaborativemapping},
which estimates per-session scale via $\mathrm{Sim}(3)$ anchor nodes~\cite{kim2010multiple}.
Each session $S_t$ maintains keyframe poses $X_i^t \in \mathrm{Sim}(3)$
in a session-local frame,
and an anchor node $A^t \in \mathrm{Sim}(3)$ maps this local frame
to the global frame,
so that the world-frame pose of keyframe $i$ is
$T_i^t = A^t \cdot X_i^t$.
Inter-session loop closures are established by matching keyframe images
across sessions using MASt3R~\cite{leroy2024grounding}
and computing relative $\mathrm{Sim}(3)$ constraints
via ray-based geometric optimization.
The resulting factor graph,
containing intra-session odometry edges and inter-session loop-closure edges,
is optimized jointly using g2o~\cite{kummerle2011g},
yielding globally consistent trajectories across all sessions.
This multi-session alignment provides the spatial backbone for LT-Mem.
Because all per-session object observations are expressed in the same
global coordinate frame,
the reasoning layer can directly compare object centroids across sessions
and detect state transitions such as displacement or disappearance.

\subsubsection{Instance Segmentation and 3D Lifting}
For each keyframe in session $S_t$,
we apply SAM3~\cite{carion2025sam}
with text prompts to obtain 2D instance masks.
Given globally consistent poses and dense pointmaps from the \ac{SLAM} backend,
each mask region is projected into the global frame
to obtain per-instance 3D points,
from which centroid $\mathbf{c}^q_t$, volume $v^q_t$,
and visual embedding $\mathbf{f}^q_t$ are computed.
Within a session, fragmented detections of the same instance are merged
using spatial consistency and point-cloud clustering,
yielding one consolidated observation $o^q_t$ per object per session.

\subsection{Reasoning Layer}
\label{sec:reasoning_layer}
\subsubsection{Cross-Session Re-Identification}
When revisiting an environment, geometric proximity alone is
insufficient for identity preservation.
Given a session-level observation and candidate tracks from previous
sessions, we compute five normalized evidence scores
(Table~\ref{tab:evidence}).
The re-identification confidence is their weighted combination,
where each $E_i \in [0,1]$ and $\sum_i w_i = 1$.

To reduce the search space, candidate tracks are pre-filtered
via embedding-based retrieval over a persistent vector store.
Top-$k$ ($k{=}5$ in all experiments) candidates are retrieved
based on cosine similarity.
If retrieval yields no confident candidates, all cross-session tracks
with compatible semantic labels are considered. Final association follows a hybrid strategy.
Simple cases are resolved by deterministic hard rules
(e.g., first observation, class mismatch).
Ambiguous cross-session cases are delegated to a constrained
LLM judge  that outputs one of
\textsc{match}, \textsc{new-track}, or \textsc{hold}
under a strictly constrained output space.
Thus, structured evidence governs most associations;
the LLM is invoked only for final disambiguation and
does not modify evidence scores, operating strictly
over pre-computed structured inputs.

\begin{table}[t!]
\centering
\caption{Evidence scores for cross-session re-identification.}
\label{tab:evidence}
\resizebox{\columnwidth}{!}{%
\begin{tabular}{c l l}
\toprule
\textbf{Score} & \textbf{Cue} & \textbf{Role} \\
\midrule
$E_1$ & Spatial proximity     & Weak prior; avoids penalizing displacement \\
$E_2$ & Temporal continuity   & Stabilizing cue across consecutive sessions \\
$E_3$ & Feature similarity    & Primary signal; robust to viewpoint change \\
$E_4$ & Motion consistency    & Conditioned on volatility estimate \\
$E_5$ & Occlusion handling    & Suppresses false disappearance \\
\bottomrule
\end{tabular}
}
\vspace{3mm}
\end{table}

%

\subsubsection{Structural Integrity Check}
\label{sec:structural_check}

Before updating memory, we verify global alignment quality
using structural anchors.
For each session, anchor centroids observed in the current session
are compared against their registered positions in Live Memory.
We compute the mean anchor displacement
\(
\bar{d}_{\text{anchor}}
\)
across all matched anchors.
If
\(
\bar{d}_{\text{anchor}} > \tau_{\text{align}},
\)
a \textsc{session hold} is triggered:
all updates for that session are skipped and an
alignment-failure event is recorded.
This mechanism prevents corrupted observations from propagating
into memory under global misalignment.

\begin{figure*}[t]
    \centering
    \includegraphics[width=\textwidth]{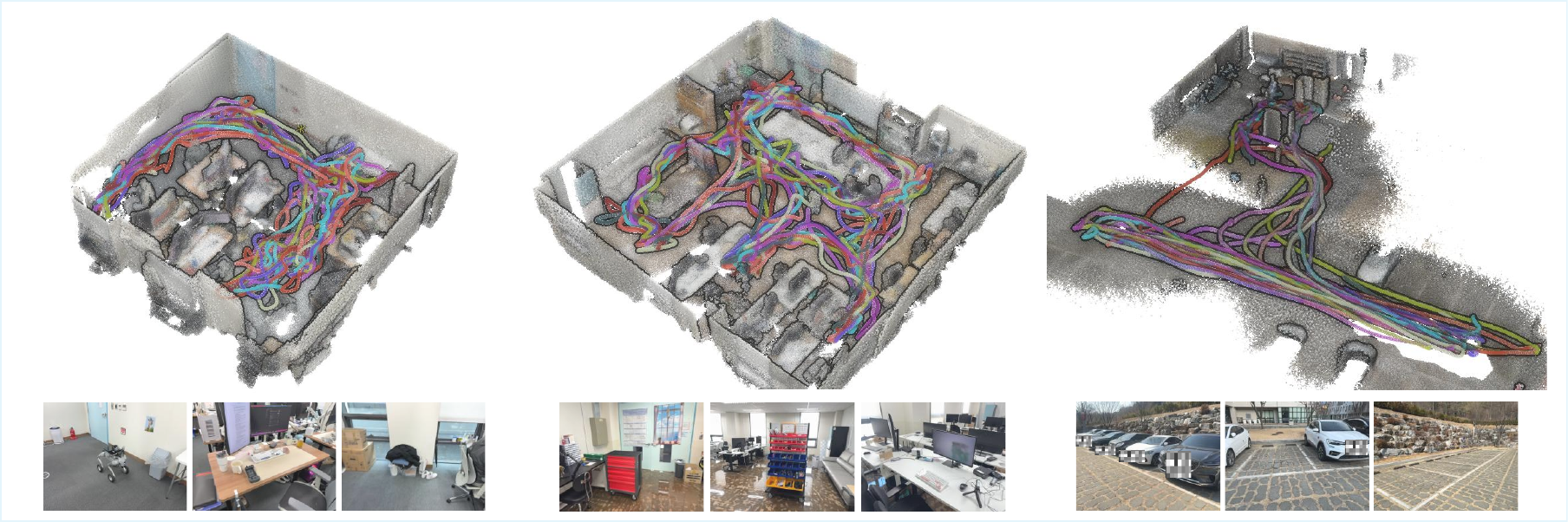}
    \caption{Overview of the three LT-VQA environments and dataset. Each column shows a globally aligned 3D point-cloud map with multi-session trajectories overlaid (top) and representative RGB frames (bottom). (Left)~\texttt{Lab-S}: a compact indoor room with 10 tracked objects. (Middle)~\texttt{Lab-L}: a larger indoor space with more complex layout. (Right)~\texttt{Parking Lot}: an outdoor scene for spatial statistics evaluation. Trajectory colors distinguish individual sessions; object configurations change across sessions due to relocation and temporary disappearance.}
    \label{fig:dataset_scenes}
    \vspace{-5mm}
\end{figure*}

\subsubsection{Change Detection}
We compute the inter-session displacement
$d = \|\mathbf{c}_t^q - \mathbf{c}_{t-1}^q\|$.
Rather than relying on a single threshold,
we apply a priority-ordered deterministic rule set that considers:
(i) alignment quality, (ii) missing-evidence signals 
(disappearance from the old location), and 
(iii) volatility-conditioned motion tolerance.

\subsubsection{Volatility-Aware Memory Update}
The volatility score $V_t \in [0,1]$ is updated using a
normalized evidence accumulation rule:
\begin{equation}
\label{eq:volatility}
  V_t =
  \frac{P(E_t \mid V_{t-1})\,V_{t-1}}
       {P(E_t \mid V_{t-1})\,V_{t-1}
       + P(E_t \mid 1{-}V_{t-1})(1{-}V_{t-1})},
\end{equation}

where $E_t \in \{\textsc{none},\allowbreak\ \textsc{move},\allowbreak\ \textsc{appear},\allowbreak\ \textsc{disappear},\allowbreak\ \textsc{re-appear}\}$
 is modeled as a fixed monotonic function of $V$
(e.g., $P(\textsc{move} \mid V)$ increases with $V$),
serving as a lightweight temporal evidence accumulator,
and is not learned. The initial prior $V_0$ is obtained via a one-time LLM query
conditioned on the object's semantic label.
Fig.~\ref{fig:volatility} illustrates how LT-Mem updates $V_t$
across sessions for objects with different dynamics;
the empirical behavior is discussed in Sec.~\ref{sec:exp}.

Given the change decision and volatility estimate,
a deterministic update policy selects one of three actions:

\begin{itemize}
\item \textsc{hold}: preserve the previous state under low confidence or occlusion,
\item \textsc{overwrite}: commit the new state,
\item \textsc{multi-hypothesis}: retain competing location hypotheses.
\end{itemize}

Under \textsc{multi-hypothesis}, a small set of candidate states
is maintained with confidence scores.
A hypothesis is promoted when its confidence exceeds competing
hypotheses by a margin $\Delta$ over consecutive sessions.
As consistent evidence accumulates for a dominant location,
competing hypotheses are consolidated.
The highest-confidence hypothesis is used for present-state queries.

\subsection{Tri-Memory Architecture}
\label{sec:trimem}

The update actions described above produce three distinct types of
outputs: a current state estimate, a timestamped event record,
and long-term statistics.
To organize these heterogeneous outputs,
memory $\mathcal{M}^q$ is structured into three complementary
components (\figref{fig:framework-overview}). We emphasize that the Tri-Memory design is not proposed as
an independent architectural contribution;
rather, it is a direct consequence of the volatility-aware
reasoning layer, which produces distinct outputs—state estimates, event records, and long-term statistics—that cannot be collapsed into a single map representation without reintroducing the
temporal amnesia described in \secref{sec:related-work}.
We note that the terminology of delta and meta maps
has precedent in lifelong mapping literature~\cite{kim2022lt},
where they denote geometric point-cloud differences and
map-level statistics used to detect and remove outdated
spatial elements.
In contrast, LT-Mem's Delta and Meta Memories operate
at the object-semantic level:
Delta Memory records identity-preserving state transitions
(e.g., a specific chair moved from A to B at session $S_5$),
while Meta Memory accumulates per-object behavioral statistics
such as volatility scores that feed back into
the update policy.
This shift from point-level map differencing to
object-level event logging is what enables
structured temporal queries that purely geometric
delta maps cannot support.

\textbf{Live Memory}
stores the current confirmed state of each object.
When the update policy selects \textsc{overwrite},
the previous state is replaced;
under \textsc{multi-hypothesis},
a small set of competing location hypotheses is maintained
alongside their confidence scores.
Present-state queries (e.g., ``Where is the white chair now?'')
are resolved directly from this component.

\textbf{Delta Memory}
records a timestamped, per-object event log.
Each update action appends a structured entry:
\textsc{move}, \textsc{appear}, \textsc{disappear},
\textsc{re-appear}, or \textsc{none},
along with session index and metric context
(e.g., displacement magnitude).
Alignment-failure events from the Structural Integrity Check
are also logged, preserving an auditable record
even when updates are suppressed.
Event-history queries (e.g., ``When did the scissors first appear?'')
and counterfactual queries
(e.g., ``Was the robot dog in the same place in S1 and S10?'')
are answered by traversing this log.

\begin{table}[t!]
\centering
\caption{LT-VQA dataset overview. (Top) Task types and target environments.
(Bottom) Recording statistics per environment.}
\label{tab:dataset}

\resizebox{\columnwidth}{!}{%
\begin{tabular}{l l l}
\toprule
\textbf{Task Type} & \textbf{Environment} & \textbf{Description} \\
\midrule
Instance History    & \texttt{Lab-S}, \texttt{Lab-L} & Per-object state transitions across sessions \\
\cmidrule(lr){1-3}
Spatial Statistics  & \texttt{Parking Lot}  & Aggregate occupancy and count patterns       \\
\bottomrule
\end{tabular}
}

\vspace{2mm}

\resizebox{\columnwidth}{!}{%
\begin{tabular}{l l c c c}
\toprule
\textbf{Env.} & \textbf{Task Type} & \textbf{\#Sess} & \textbf{Frames/Sess} & \textbf{Duration/Sess} \\
\midrule
\texttt{Lab-S}       & Instance History   & 10 & 116--159 & 1.6--2.1 min \\
\texttt{Lab-L}       & Instance History   & 10 & 115--183 & 2.6--3.3 min \\
\cmidrule(lr){1-5}
\texttt{Parking Lot} & Spatial Statistics & 10 & 124--195 & 2.2--3.2 min \\
\bottomrule

\vspace{2mm}

\end{tabular}

}

\vspace{-4mm}
\end{table}

\textbf{Meta Memory}
accumulates long-term statistics derived from the event log,
including the volatility score $V_t$ and per-object change frequency.
These statistics feed back into the reasoning layer:
$V_t$ conditions the motion tolerance in subsequent sessions,
closing the loop between observation and update policy.
Aggregate queries
(e.g., ``Which object moved most frequently?'')
are answered from this component.

At query time, an object-centric retrieval module
routes each question to the relevant memory component(s)
based on query type (\tabref{tab:transition_taxonomy}).
Metric queries such as displacement or frequency
are answered via deterministic computations over the event log,
ensuring numerically grounded responses
independent of LLM hallucination.

\section{Experiments}
\label{sec:exp}

\subsection{Dataset}
\label{subsec:dataset}
We construct LT-VQA, a controlled multi-session dataset 
for evaluating long-term object-centric scene understanding.
It jointly provides globally aligned multi-session geometry, 
persistent object-instance identity, event-level state 
transition annotations, and session-indexed temporal QA 
pairs. While 3RScan ~\cite{wald2019rio} provides 
instance-level correspondence across rescans, it lacks 
event-level annotations and temporal QA supervision. To our knowledge, no existing multi-session dataset jointly provides these components; unlike single-session VQA, cross-session event annotation requires meticulous multi-temporal alignment per object, making dense labeling inherently resource-intensive.

\begin{figure}[t]
\centering
\includegraphics[width=\columnwidth]{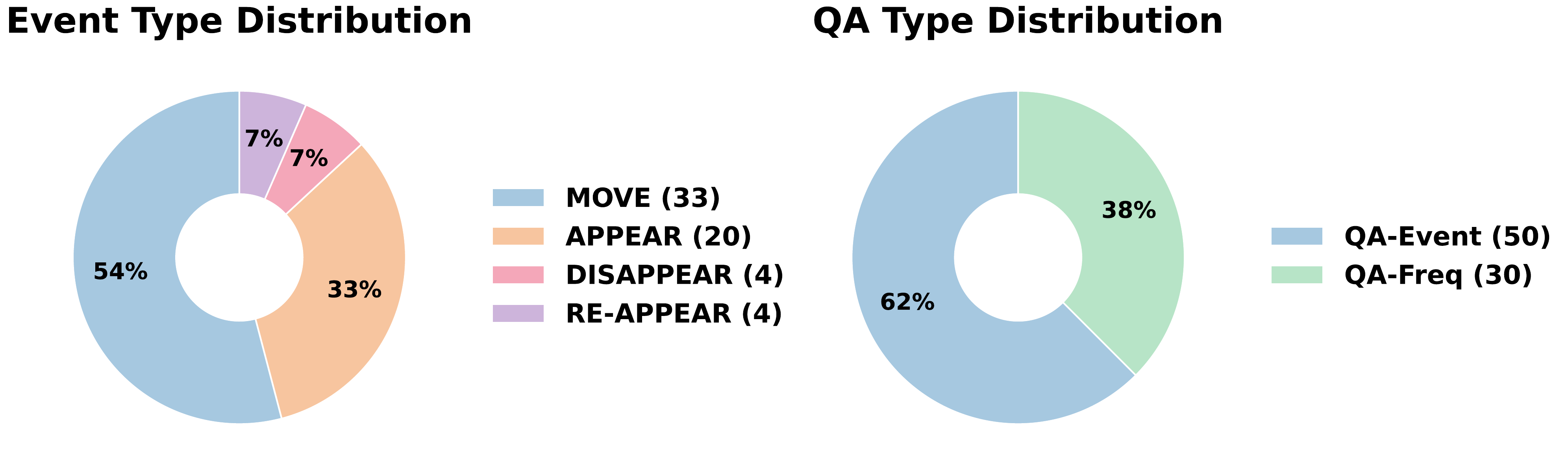}
\caption{LT-VQA annotation statistics.
(Left) Ground-truth event type distribution across
Lab-S and Lab-L (61  state-change events; NONE events omitted).
(Right) QA pair distribution by query type (80 pairs total).}
\label{fig:dataset-dist}
\end{figure}

Two indoor environments, \texttt{Lab-S} and \texttt{Lab-L},
each track 10 objects over 10 sessions with evolving object configurations across sessions (Table~\ref{tab:dataset}),
ranging from highly volatile instances (e.g., brown basket, scissors) to fully stationary ones (e.g., fridge, sofa). Representative queries for each event type are listed in Table~\ref{tab:transition_taxonomy}.
Ground-truth annotations provide per-session event labels (\textsc{appear}, \textsc{disappear}, \textsc{move}, 
\textsc{re-appear}, \textsc{none}); annotation statistics are shown in
Fig.~\ref{fig:dataset-dist}.
The \texttt{Parking Lot} sequence extends evaluation to an outdoor
setting with 10 sessions recorded across different days and times of day, targeting aggregate 
spatio-temporal patterns.
\subsection{Experimental Setup}
\label{subsec:setup}
\noindent \textbf{Data Collection.}
All sequences are captured as monocular RGB video (Lab-L and Parking Lot with iPhone 15 Pro; Lab-S with iPhone 17 Pro) and processed through our multi-session 
\ac{SLAM} pipeline.

\noindent  \textbf{Implementation Details.}
All experiments are conducted on an NVIDIA RTX 5070 Ti GPU
with 64\,GB of RAM. For LLM-dependent components, both LT-Mem and applicable 
baselines use Gemini~2.5~Pro to ensure consistent comparison.
Object-centric retrieval uses BGE-small-en-v1.5 with
ChromaDB, without invoking the LLM during retrieval. For cross-session re-identification (Table~\ref{tab:evidence}), the evidence weights are $w_1{=}0.05$, $w_2{=}0.20$, $w_3{=}0.45$, $w_4{=}0.15$, $w_5{=}0.15$. The alignment gating threshold (Sec.~\ref{sec:structural_check})  is $\tau_{\text{align}}{=}0.3$\,m.

\subsection{Evaluation Protocol}
\label{subsec:protocol}
%
\begin{figure}[t!]
  \centering
  \includegraphics[width=\columnwidth]{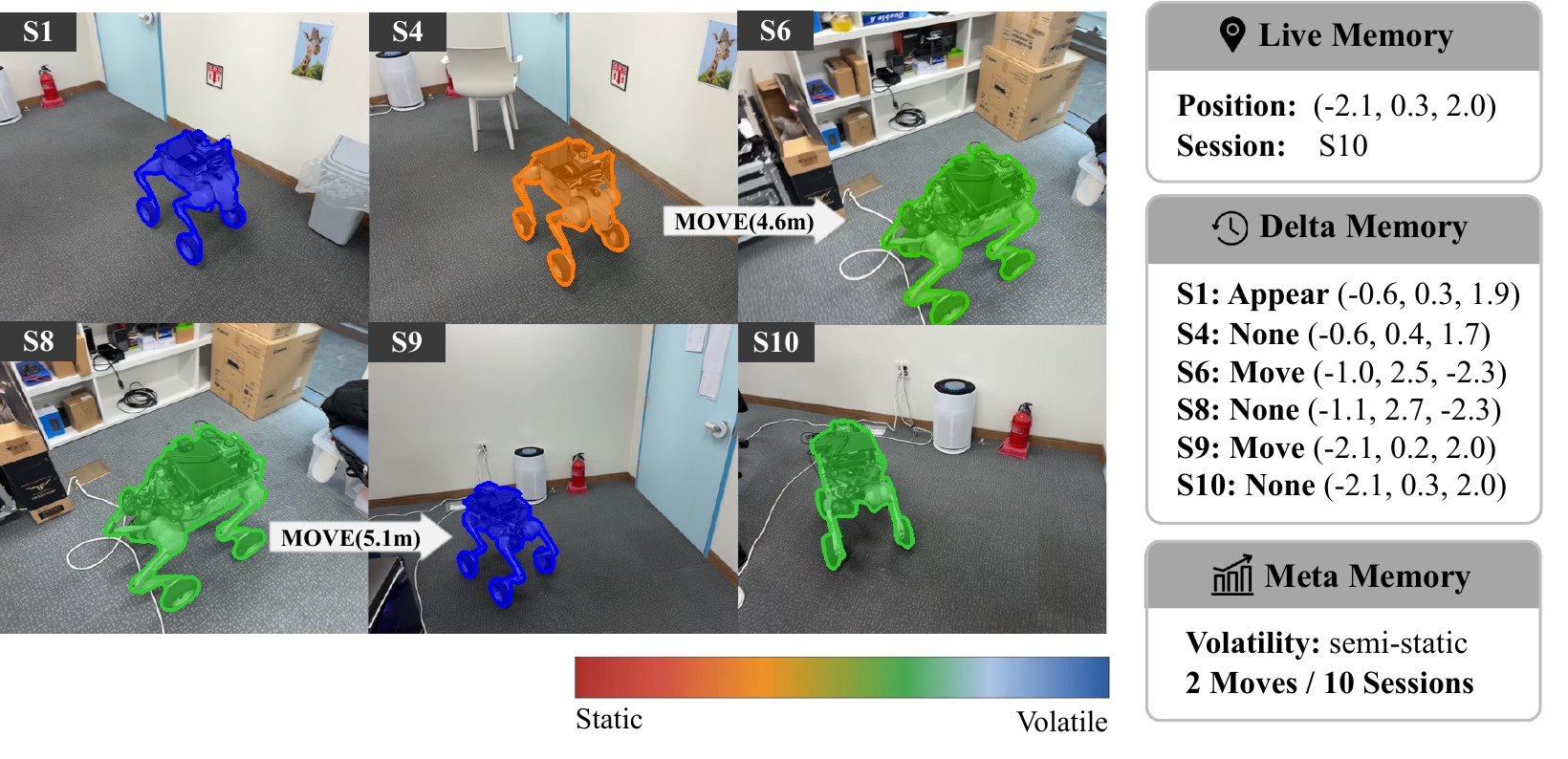}
    \caption{Tri-Memory example for the robot dog in \texttt{Lab-S}
over 10 sessions. The color overlay encodes the evolving
volatility score $V_t$ (red: static, blue: volatile).
Two \textsc{move} events are recorded
in Delta Memory with metric context; Live Memory reflects the
final confirmed position at S10, Meta Memory classifies the
object as semi-static.}
  \label{fig:qual}
\end{figure}

\begin{figure}[t!]
  \centering
  \includegraphics[width=\linewidth,height=0.18\textheight,keepaspectratio]{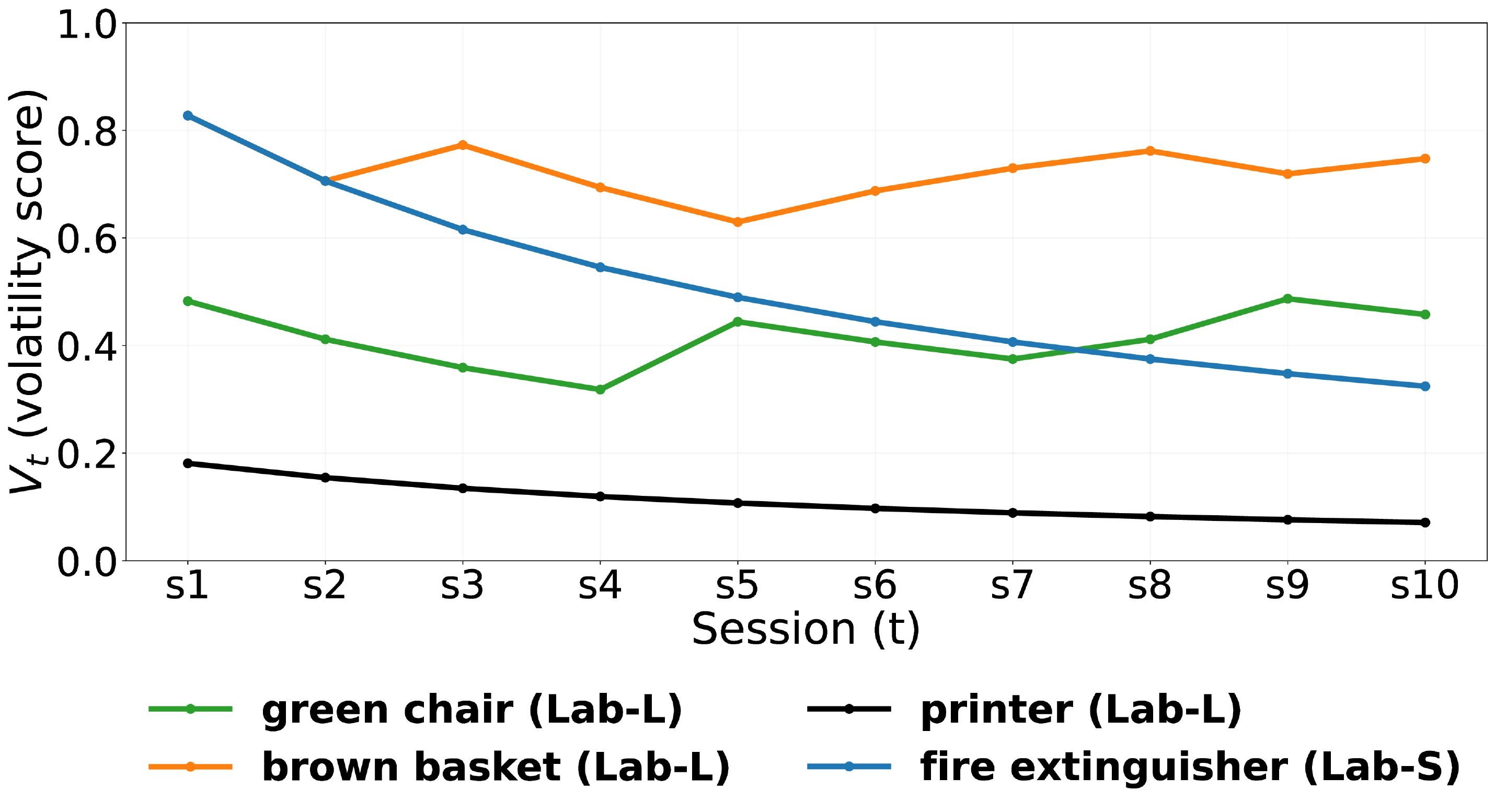}
    \caption{Evolution of $V_t$ across sessions ($t=0$: LLM-initialized
prior). Four representative dynamics are shown: persistent
high volatility (brown basket); progressive correction of an
initially overestimated prior (fire extinguisher); mid-range
fluctuation with partial recovery (green chair); and near-zero
convergence throughout (printer).}
  \label{fig:volatility}
\end{figure}

%

%

We report three metrics for Instance History Tracking.
An \emph{event} denotes an object-level state transition
between consecutive sessions---one of
\textsc{move}, \textsc{appear}, \textsc{disappear}, \textsc{re-appear}, or \textsc{none}.
Event~F1 treats each ground-truth event as a positive instance,
measuring detection precision and recall.
QA-Event measures exact-match accuracy on natural-language
queries about object event histories
(e.g., \textit{``What happened to the robot dog at Session~6?''}).
QA-Freq evaluates frequency queries
(e.g., \textit{``How many times did the robot dog move?''})
against ground-truth statistics accumulated in Meta Memory. QA-Event and QA-Freq are evaluated automatically by matching structured JSON outputs against ground-truth annotations.

For Instance History Tracking, we compare against four baselines that represent distinct paradigms for multi-session object reasoning. Existing systems such as Khronos ~\cite{schmid2024khronos}, ConceptGraphs ~\cite{gu2024conceptgraphs}, and ReMEmbR ~\cite{anwar2025remembr} do not produce object-level event logs or support temporal QA, and augmenting each to do so would require modules outside their original design scope. Our baselines instead isolate core reasoning paradigms to enable controlled evaluation.

\noindent\textbf{Geometric Only.}
Following the geometry-centric lifelong mapping
paradigm~\cite{kim2022lt}, we construct a baseline that detects
object-level changes via distance thresholding over session
observations. Because this approach maintains only the current
map state without an event log, event-history and frequency
queries are structurally unanswerable, reported as N/A.

\begin{table}[t]
\centering
\caption{Representative QA outputs from LT-Mem, grouped by the memory component used for retrieval.}
\label{tab:qa_examples}
\setlength{\tabcolsep}{3pt}
\renewcommand{\arraystretch}{1.15}
\footnotesize
\begin{tabularx}{\columnwidth}{>{\raggedright\arraybackslash}p{0.95\columnwidth}}
\toprule
\multicolumn{1}{c}{\textbf{Delta Memory}} \\
\midrule
\textit{Q: ``When did the scissors first appear?''} \\
A: The scissors first appeared in Session 3. \\[3pt]
\textit{Q: ``What happened to the robot dog at Session 6?''} \\
A: At Session 6, the robot dog moved from $(-0.81,\; 0.43,\; 1.77)$ to $(-0.97,\; 2.52,\; -2.27)$. \\
\midrule
\multicolumn{1}{c}{\textbf{Live Memory}} \\
\midrule
\textit{Q: ``Where did the trash bin end up after all sessions?''} \\
A: At the end (S10), the trash bin was at $(-0.65,\; 0.52,\; 1.49)$. \\[3pt]
\textit{Q: ``Where is the green chair now?''} \\
A: At $(2.95,\; 0.89,\; 0.25)$ as of Session 10. \\
\midrule
\multicolumn{1}{c}{\textbf{Meta Memory}} \\
\midrule
\textit{Q: ``Can I use the printer as a navigation landmark?''} \\
A: Yes. Volatility 0.02, highly stable. Use its position at $(-1.75,\; 0.28,\; 0.50)$ from S1 as reference. \\[3pt]
\textit{Q: ``Which object moved most frequently?''} \\
A: The brown basket (4 moves). \\
\bottomrule
\end{tabularx}
\end{table}

\noindent\textbf{Text-Batch.}
Each keyframe is independently captioned and aggregated into
a session-level summary. Object-level events are inferred
by comparing consecutive session summaries.
Without explicit cross-session object identity or spatial
grounding, this approach cannot reliably detect spatially
grounded transitions such as \textsc{move} and
\textsc{re-appear}; only coarse presence changes
(\textsc{appear}, \textsc{disappear}) are partially captured.

\noindent\textbf{VLM-Batch.}
For each object, all keyframes across all sessions are provided as a single batch query for direct state-transition judgment.
This baseline has full temporal visual context but relies
entirely on the VLM's implicit reasoning without structured
memory or volatility-aware update logic.

\noindent\textbf{STAR.}
We evaluate STAR~\cite{chen2026searchingspacetimeunified}, a recent agentic memory--action retrieval system, directly on LT-VQA, disabling only its embodied navigation actions---inapplicable in our passive setting---so the agent answers each query by iteratively retrieving captioned and raw visual records from its non-parametric memory.

Spatial Statistics targets aggregate scene-level patterns rather than per-object tracking; we evaluate this capability qualitatively with occupancy analysis across sessions.

\subsection{Result Overview and Tri-Memory Examples}
\label{subsec:result_overview}

%
\begin{figure}[t] 
\centering

\begin{minipage}{1.0\columnwidth}
    \centering
    \captionsetup{type=table}
    \captionof{table}{Main results on LT-VQA (\texttt{Lab-S} and \texttt{Lab-L} combined). LT-Mem outperforms baselines across all metrics with high token efficiency.}
    \label{tab:main}
    \vspace{0.5em}
    \resizebox{\columnwidth}{!}{
    \begin{tabular}{l | c c c | c}
    \toprule
    \textbf{Method} & \textbf{Event F1} $\uparrow$ & \textbf{QA-Event} $\uparrow$ & \textbf{QA-Freq} $\uparrow$ & \textbf{Tokens} $\downarrow$ \\
    \midrule
    Geometric Only & 0.630 & N/A   & N/A   & -- \\
    Text-Batch     & 0.460 & 0.300 & 0.267 & 3,973K \\
    VLM-Batch      & 0.790 & 0.680 & 0.333 & 7,114K \\
    STAR$^\dagger$ \cite{chen2026searchingspacetimeunified}& 0.420 & 0.460 & N/A & 45,859K \\
    \midrule
    \textbf{Ours} (Qwen2.5)    & 0.885 & 0.800 & 0.567 & \textbf{352K} \\
    \textbf{Ours} (Gemini 2.5) & \textbf{0.910} & \textbf{0.820} & \textbf{0.600} & 438K \\
    \bottomrule
    \end{tabular}%
    }
    
    \vspace{2pt}
    {\footnotesize $^\dagger$STAR emits free-form text, so its QA-Event is scored by keyword match; all other methods use strict exact-match.\par}
    \label{tab:main_tab}
\end{minipage}

\vspace{3mm} 

\begin{minipage}{1.0\columnwidth}
    \centering
        \includegraphics[width=\textwidth]{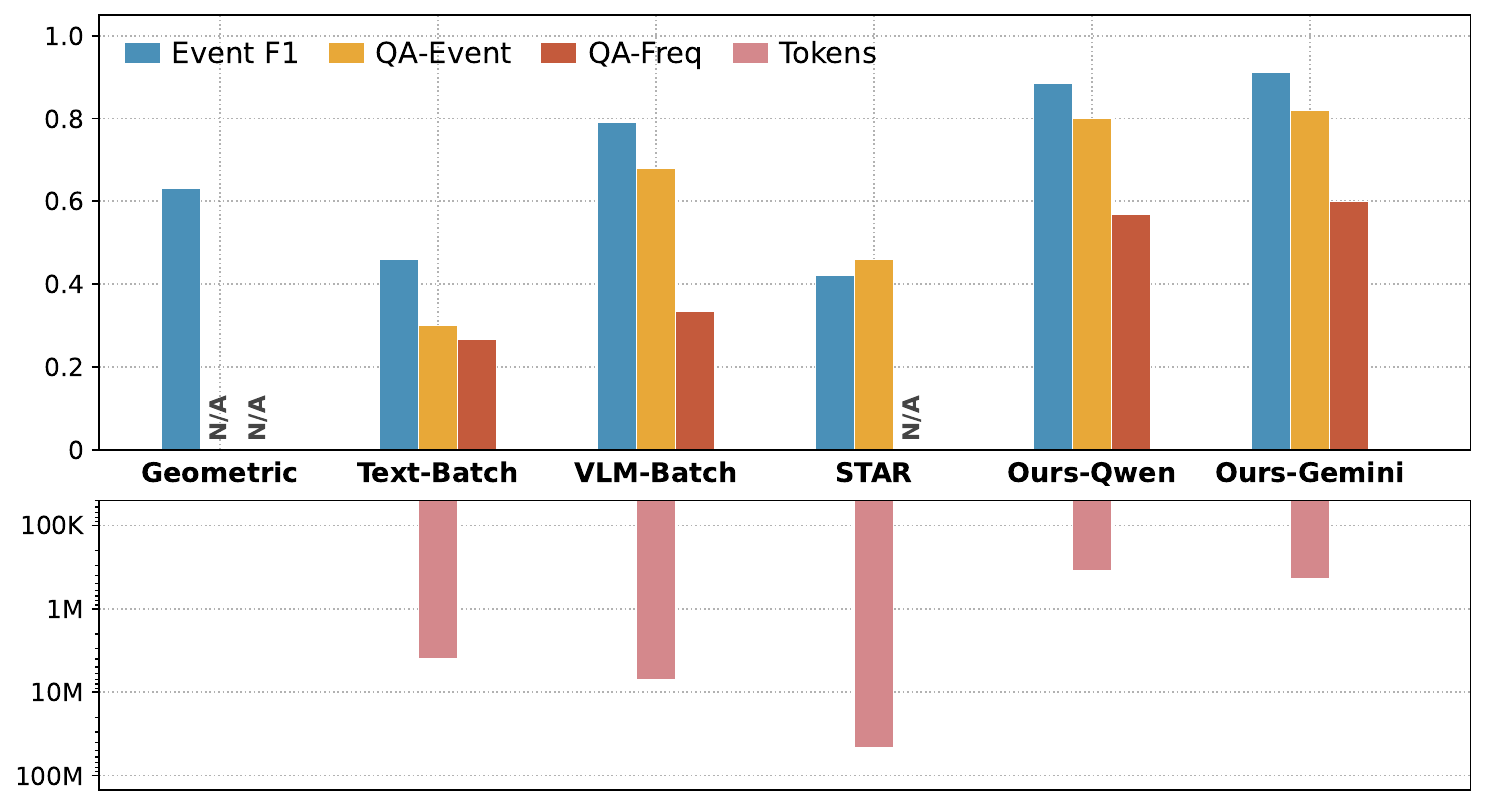}
    \captionof{figure}{Visualization of the quantitative results. {(Top)~Scores across three evaluation metrics.
(Bottom)~Cumulative LLM/VLM API tokens over the evaluation set (log scale).
LT-Mem achieves the highest scores across all metrics
while consuming far fewer tokens than vision-based baselines.}}
    \label{fig:main_vis}
\end{minipage}

\end{figure}

We illustrate LT-Mem's behavior through qualitative examples
across the three memory components.
Fig.~\ref{fig:qual} illustrates the volatility-aware update
process for the robot dog in \texttt{Lab-S}.
The volatility score $V_t$ increases as MOVE events accumulate and decreases during stationary periods.
Two relocations at S6 and S9 are committed as \textsc{move}
events, while small displacements at S4, S8, and S10 fall below the volatility-conditioned motion tolerance and are recorded as \textsc{none} events in Delta Memory.
Fig.~\ref{fig:volatility} illustrates how $V_t$ adapts across sessions for objects with different dynamics. 
Even when the LLM-initialized prior ($t=0$) deviates from
an object's actual dynamics, $V_t$ converges within a few
sessions through the Bayesian update (Eq.~\ref{eq:volatility}),
removing the need for per-object tuning.
This is important because an object's volatility is not
a fixed class-level property---the same object type may
exhibit different dynamics depending on the environment
and usage context.
Table~\ref{tab:qa_examples} presents representative QA outputs
grouped by memory component.
The examples show that different temporal queries require
access to different aspects of object history,
and that collapsing any memory component would make
the corresponding query type unanswerable.

\subsection{Instance History Tracking}
\label{subsec:instance_history}

Table~\ref{tab:main} reports results on \texttt{Lab-S} and \texttt{Lab-L} combined. The geometry-only baseline achieves reasonable Event~F1 through distance thresholding, but cannot answer any temporal query without an event log (N/A)---a direct consequence of temporal amnesia. 
Text-Batch achieves low scores across all metrics, confirming that
frame-level captioning without spatial grounding or
cross-session identity is insufficient for temporal reasoning.
VLM-Batch achieves competitive scores with full visual
context per object, but at substantially higher token cost. STAR achieves the lowest Event F1 despite the highest token consumption: without persistent identity or an event log, its per-query retrieval over raw records cannot compose coherent object histories, and frequency queries requiring structured counts are unanswerable (N/A).

Token counts in Table~\ref{tab:main} reflect cumulative runtime token usage over the evaluation set; perception (segmentation, feature
extraction) is a one-time offline cost amortized over all
subsequent queries.
Counts for Gemini~2.5~Pro and Qwen2.5-3B reflect each
model's native tokenizer; despite different absolute values,
both remain well below the vision-based baselines, which
consume far more tokens by repeatedly feeding raw frames or
visual records to the model.
LT-Mem requires far fewer tokens by converting visual
observations into structured representations during perception,
so that the reasoning layer receives only compact textual inputs.
Because LT-Mem's per-session token budget is independent of
frame count, this gap widens as sessions grow---a key advantage
for lifelong operation.

\textbf{LT-Mem (Full)} consistently improves over all baselines across
applicable metrics, even when using Qwen2.5-3B as a lightweight
local LLM, indicating that
the gain stems from the structured memory architecture rather
than LLM capacity.
\begin{figure}[t!]
  \centering
  \includegraphics[width=\linewidth,keepaspectratio]{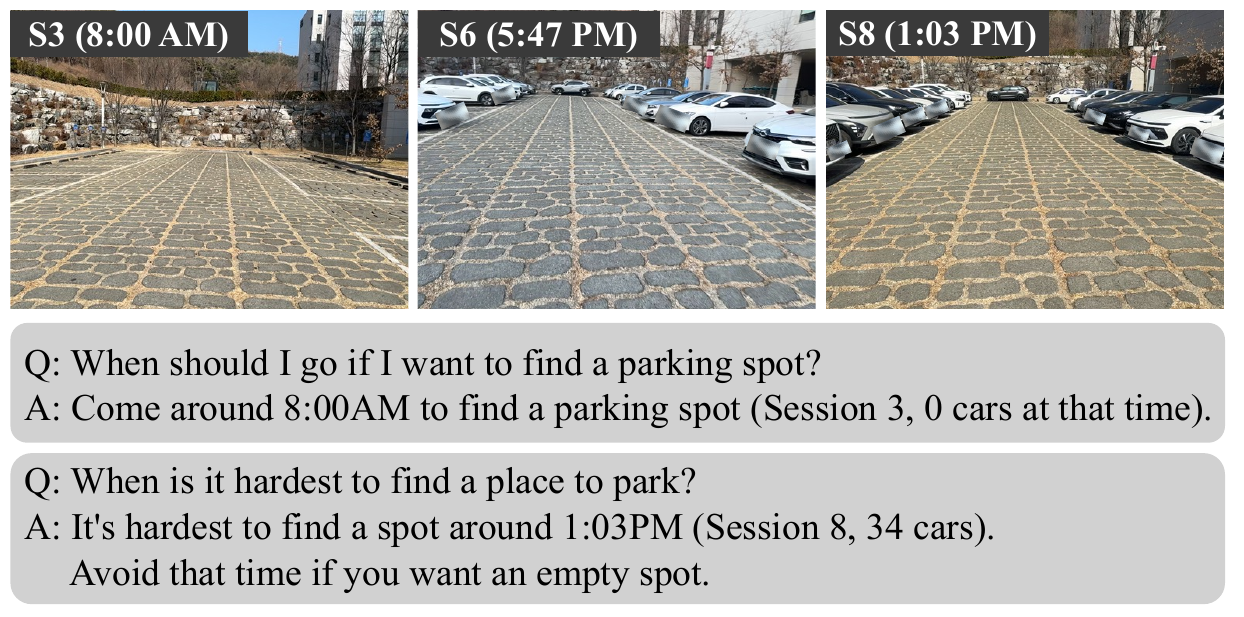}
  \caption{Spatial Statistics on the \texttt{Parking Lot} sequence.
Three sessions captured on different days at varying times show varying occupancy levels.
Meta Memory aggregates per-session vehicle counts,
enabling deterministic answers to temporal occupancy queries
without additional inference.}
\label{fig:spatial-stats}
\vspace{-2mm}
\end{figure}

%

\subsection{Spatial Statistics}
\label{subsec:spatial_stats}

The \texttt{Parking Lot} sequence evaluates Spatial Statistics,
a task type structurally distinct from Instance History Tracking:
rather than tracking specific instances, the system summarizes
session-level occupancy patterns through Meta Memory.
Fig.~\ref{fig:spatial-stats} shows occupancy trends across
sessions alongside example queries such as
\textit{``When should I go if I want to find a parking spot?''}
and \textit{``When is it hardest to find a place to park?''}
By aggregating per-session object counts in Meta Memory,
LT-Mem answers these queries through deterministic lookup
over stored statistics without additional inference.
\begin{table}[t!]
\centering
\caption{Ablation study on LT-VQA (\texttt{Lab-S} and \texttt{Lab-L} combined).}
\label{tab:ablation}
\resizebox{0.85\columnwidth}{!}{%
\setlength{\tabcolsep}{8pt}
\begin{tabular}{l || c | c | c}
\toprule
\textbf{Method}
  & \textbf{Event F1} $\uparrow$
  & \textbf{QA-Event} $\uparrow$
  & \textbf{QA-Freq} $\uparrow$ \\
\midrule
w/o Re-ID        & 0.140 & 0.120 & 0.070 \\
w/o Volatility   & 0.615 & 0.640 & 0.130 \\
\midrule
\textbf{Ours (Full)} & \textbf{0.910} & \textbf{0.820} & \textbf{0.600} \\
\bottomrule
\end{tabular}
}%
\end{table}

%
\subsection{Ablation Study}
\label{subsec:ablation}

Table~\ref{tab:ablation} presents component-level ablation results
on \texttt{Lab-S} and \texttt{Lab-L} combined.
\textbf{Structural Integrity Check:} Mean anchor displacement remained below the gating threshold
throughout all experiments; no session was rejected. This confirms that the multi-session SLAM alignment (Sec.~III-B) provides sufficiently accurate global registration for downstream reasoning.
\textbf{Effect of Re-Identification:} Removing cross-session identity matching and assigning each observation as an independent track (w/o Re-ID) causes
near-complete collapse across all metrics. This confirms that persistent identity is the foundational prerequisite for all downstream temporal reasoning---without it, Delta Memory cannot record identity-preserving transitions and Meta Memory loses the continuity required for meaningful statistics.

\textbf{Effect of Volatility-Aware Update:} Replacing volatility-conditioned updates with a fixed threshold (w/o Volatility) degrades all metrics, as the fixed threshold cannot adapt to per-object dynamics, producing false events that corrupt Meta Memory statistics.

\section{Conclusion}
\label{sec:conclusion}
We introduced LT-Mem, a volatility-aware memory evolution framework that unifies multi-session spatial alignment with volatility-conditioned update logic for lifelong robot operation.
The resulting Tri-Memory structure (Live, Delta, Meta) preserves both current states and event histories, enabling longitudinal object-centric reasoning that cannot be achieved by maintaining only a single current map state.
Experiments on LT-VQA validate the effectiveness of structured memory evolution for preserving object histories and supporting temporal queries across multi-session revisits with practical token efficiency that scales to longer deployments.
While LT-VQA currently comprises two indoor and one outdoor environment over 10 sessions, broader validation would further assess generalization.
Cross-session re-identification also remains challenging when multiple objects share similar appearance and spatial context, reflecting a fundamental difficulty in identity preservation under ambiguity.
Scaling to longer deployment horizons and more diverse environments, with the goal of establishing LT-VQA as a public benchmark for longitudinal object-centric reasoning, is a promising direction for future work.


\bibliographystyle{unsrtnat}   

{\footnotesize
\bibliography{string-short, references}

@string{C-ICRA = {Proc. IEEE Int. Conf. Robot. Autom.}}

@string{C-CVPR = {Proc. IEEE/CVF Conf. Comput. Vis. Pattern Recognit.}}

@string{C-ICCV = {Proc. IEEE/CVF Int. Conf. Comput. Vis.}}

@string{C-ECCV = {Proc. Eur. Conf. Comput. Vis.}}

@string{C-RSS  = {Proc. Robot. Sci. Syst.}}

@inproceedings{kim2022lt,
  title={LT-mapper: A Modular Framework for {LiDAR}-Based Lifelong Mapping},
  author={Kim, Giseop and Kim, Ayoung},
  booktitle=C-ICRA,
  pages={7995--8002},
  year={2022}
}

@book{carlone2025slam,
  title={{SLAM} Handbook: From Localization and Mapping to Spatial Intelligence},
  author={Carlone, Luca and Kim, Ayoung and Barfoot, Timothy and Cremers, Daniel and Dellaert, Frank},
  year={2025},
  publisher={Cambridge University Press}
}

@inproceedings{murai2025mast3r,
  title={{MASt3R-SLAM}: Real-Time Dense {SLAM} with {3D} Reconstruction Priors},
  author={Murai, Riku and Dexheimer, Eric and Davison, Andrew J},
  booktitle=C-CVPR,
  pages={16695--16705},
  year={2025}
}

@article{cadena2017past,
  title={Past, Present, and Future of Simultaneous Localization and Mapping: Toward the Robust-Perception Age},
  author={Cadena, Cesar and Carlone, Luca and Carrillo, Henry and Latif, Yasir and Scaramuzza, Davide and Neira, Jos{\'e} and Reid, Ian and Leonard, John J},
  journal=IEEE_J_RO,
  volume={32},
  number={6},
  pages={1309--1332},
  year={2016}
}

@article{yin2025general,
  title={General Place Recognition Survey: Towards Real-World Autonomy},
  author={Yin, Peng and Jiao, Jianhao and Zhao, Shiqi and Xu, Lingyun and Huang, Guoquan and Choset, Howie and Scherer, Sebastian and Han, Jianda},
  journal=IEEE_J_RO,
  year={2025}
}

@inproceedings{schmid2024khronos,
  title={Khronos: A Unified Approach for Spatio-Temporal Metric-Semantic {SLAM} in Dynamic Environments},
  author={Schmid, Lukas and Abate, Marcus and Chang, Yun and Carlone, Luca},
  booktitle=C-RSS,
  year={2024}
}

@inproceedings{pomerleau2014long,
  title={Long-Term {3D} Map Maintenance in Dynamic Environments},
  author={Pomerleau, Fran{\c{c}}ois and Kr{\"u}si, Philipp and Colas, Francis and Furgale, Paul and Siegwart, Roland},
  booktitle=C-ICRA,
  pages={3712--3719},
  year={2014}
}

@article{carion2025sam,
  title={{SAM} 3: Segment Anything with Concepts},
  author={Carion, Nicolas and Gustafson, Laura and Hu, Yuan-Ting and Debnath, Shoubhik and Hu, Ronghang and Suris, Didac and Ryali, Chaitanya and Alwala, Kalyan Vasudev and Khedr, Haitham and Huang, Andrew and others},
  journal={arXiv preprint arXiv:2511.16719},
  year={2025}
}

@inproceedings{kim2010multiple,
  title={Multiple Relative Pose Graphs for Robust Cooperative Mapping},
  author={Kim, Been and Kaess, Michael and Fletcher, Luke and Leonard, John and Bachrach, Abraham and Roy, Nicholas and Teller, Seth},
  booktitle=C-ICRA,
  pages={3185--3192},
  year={2010}
}

@inproceedings{kummerle2011g,
  title={g2o: A General Framework for Graph Optimization},
  author={K{\"u}mmerle, Rainer and Grisetti, Giorgio and Strasdat, Hauke and Konolige, Kurt and Burgard, Wolfram},
  booktitle=C-ICRA,
  pages={3607--3613},
  year={2011}
}

@inproceedings{leroy2024grounding,
  title={Grounding Image Matching in {3D} with {MASt3R}},
  author={Leroy, Vincent and Cabon, Yohann and Revaud, J{\'e}r{\^o}me},
  booktitle=C-ECCV,
  pages={71--91},
  year={2024}
}

@inproceedings{anwar2025remembr,
  title={Remembr: Building and Reasoning over Long-Horizon Spatio-Temporal Memory for Robot Navigation},
  author={Anwar, Abrar and Welsh, John and Biswas, Joydeep and Pouya, Soha and Chang, Yan},
  booktitle=C-ICRA,
  pages={2838--2845},
  year={2025}
}

@inproceedings{yang20253d,
  title={{3D-Mem}: {3D} Scene Memory for Embodied Exploration and Reasoning},
  author={Yang, Yuncong and Yang, Han and Zhou, Jiachen and Chen, Peihao and Zhang, Hongxin and Du, Yilun and Gan, Chuang},
  booktitle=C-CVPR,
  pages={17294--17303},
  year={2025}
}

@inproceedings{gu2024conceptgraphs,
  title={ConceptGraphs: Open-Vocabulary {3D} Scene Graphs for Perception and Planning},
  author={Gu, Qiao and Kuwajerwala, Ali and Morin, Sacha and Jatavallabhula, Krishna Murthy and Sen, Bipasha and Agarwal, Aditya and Rivera, Corban and Paul, William and Ellis, Kirsty and Chellappa, Rama and others},
  booktitle=C-ICRA,
  pages={5021--5028},
  year={2024}
}

@inproceedings{wang2024dust3r,
  title={{DUSt3R}: Geometric {3D} Vision Made Easy},
  author={Wang, Shuzhe and Leroy, Vincent and Cabon, Yohann and Chidlovskii, Boris and Revaud, Jerome},
  booktitle=C-CVPR,
  pages={20697--20709},
  year={2024}
}

@article{qian2022pocd,
  title={{POCD}: Probabilistic Object-Level Change Detection and Volumetric Mapping in Semi-Static Scenes},
  author={Qian, Jingxing and Chatrath, Veronica and Yang, Jun and Servos, James and Schoellig, Angela P and Waslander, Steven L},
  journal={arXiv preprint arXiv:2205.01202},
  year={2022}
}

@inproceedings{wald2019rio,
  title={{RIO}: {3D} Object Instance Re-Localization in Changing Indoor Environments},
  author={Wald, Johanna and Avetisyan, Armen and Navab, Nassir and Tombari, Federico and Nie{\ss}ner, Matthias},
  booktitle=C-ICCV,
  pages={7658--7667},
  year={2019}
}

@article{gorlo2025describe,
  title={Describe Anything Anywhere At Any Moment},
  author={Gorlo, Nicolas and Schmid, Lukas and Carlone, Luca},
  journal={arXiv preprint arXiv:2512.00565},
  year={2025}
}

@article{ju2026mrscalemasterscaleconsistentcollaborativemapping,
  title={{MR.ScaleMaster}: Scale-Consistent Collaborative Mapping from Crowd-Sourced Monocular Videos},
  author={Ju, Hyoseok and Kim, Giseop},
  journal={arXiv preprint arXiv:2604.11372},
  year={2026}
}

@inproceedings{chen2026searchingspacetimeunified,
  title={Searching in Space and Time: Unified Memory-Action Loops for Open-World Object Retrieval},
  author={Chen, Taijing and Kumar, Sateesh and Xu, Junhong and Pavlakos, Georgios and Biswas, Joydeep and Mart{\'i}n-Mart{\'i}n, Roberto},
  booktitle=C-ICRA,
  year={2026}
}

@STRING{C-CVPR = {Proc. {IEEE} Conf. on Comput. Vision and Pattern Recog.}}

@STRING{C-ECCV = {Proc. European Conf. on Comput. Vision}}

@STRING{C-ICCV = {Proc. {IEEE} Intl. Conf. on Comput. Vision}}

@STRING{C-ICRA = {Proc. {IEEE} Intl. Conf. on Robot. and Automat.}}

@STRING{C-RSS = {Proc. Robot.: Science \& Sys. Conf.}}

@STRING{IEEE_J_RO = {{IEEE} Trans. Robot.}}
}

\end{document}